\documentclass[english]{article}
\usepackage{hyperref}

\usepackage[accepted]{icml2015_arxiv_mod}
\usepackage{url}
\usepackage{graphicx}
\usepackage{epsfig,subfigure}
\usepackage{amssymb}
\usepackage{amsthm}
\usepackage{amsfonts}
\usepackage{amsmath}
\usepackage{natbib}
\usepackage{babel}
\usepackage{algorithm}
\usepackage{algorithmic}
\usepackage{multicol}
\usepackage{multirow}
\usepackage{color,soul}
  \theoremstyle{plain}
  \newtheorem{lem}{\protect\lemmaname}
  \providecommand{\lemmaname}{Lemma}

\newcommand{\vertiii}[1]{{\left\vert\kern-0.25ex\left\vert\kern-0.25ex\left\vert #1\right\vert\kern-0.25ex\right\vert\kern-0.25ex\right\vert}}

\usepackage[disable]{todonotes}

\begin{document}


\twocolumn[
\icmltitle{Unbiased Bayes for Big Data: Paths of Partial Posteriors}
\icmlauthor{Heiko Strathmann${}^\star$}{heiko.strathmann@gmail.com}
\icmlauthor{Dino Sejdinovic${}^\dagger$}{dino.sejdinovic@gmail.com}
\icmlauthor{Mark Girolami${}^+$}{m.girolami@warwick.ac.uk}
\icmladdress{${}^\star$Gatsby Unit, University College London, UK\\
${}^\dagger$Department of Statistics, University of Oxford, UK\\
${}^+$Department of Statistics, University of Warwick, UK}
\icmlkeywords{MCMC, Big Data, Debiasing, Bayesian}
]

\begin{abstract}


A key quantity of interest in Bayesian inference are expectations of functions with respect to a posterior distribution.
Markov Chain Monte Carlo is a fundamental tool to consistently compute these expectations via averaging samples drawn from an approximate posterior.
 However, its feasibility is being challenged in the era of so called Big Data as all data needs to be processed in every iteration.
Realising that such \emph{simulation} is an unnecessarily hard problem if the goal is \emph{estimation}, we construct a computationally scalable methodology that allows unbiased estimation of the required expectations --  without explicit simulation from the full posterior.
The scheme's variance is finite by construction and straightforward to control, leading to algorithms that are provably unbiased and naturally arrive at a desired error tolerance.
This is achieved at an average computational complexity that is \emph{sub-linear} in the size of the dataset and its free parameters are easy to tune.
We demonstrate the utility and generality of the methodology on a range of common statistical models applied to large-scale benchmark and real-world datasets.

\end{abstract}

\listoftodos

\todo{Balaji: choice of functions: can you use this idea for estimation of marginal likelihood? (although maybe MCMC is not suitable for this) bayesians care about marginal likelihood since it's necessary for proper model comparison ... would be nice to address this problem. }

\section{Introduction}
\label{sec:Intro}

Markov Chain Monte Carlo (MCMC), used for sampling from posterior distributions, is one of the most fundamental tools in Bayesian data analysis. However, the recent explosion in the amount of data to be analysed poses serious challenges for this methodology as it is often infeasible to scale it to today's statistical problems. This development led to a recent focus on methods to make
MCMC practical for large datasets. 
Most research thus far has focused on devising alternative Markov transition kernels based on \emph{mini-batches} of the data. These approaches either lead to (1) samples being drawn from an asymptotically approximate posterior distribution, thus reducing the amount of computation at the expense of introducing bias \citep{Welling2011,Korattikara2014,ChenFoxGuestrin2014,Bardenet2014}, or (2) preserving the asymptotically correct invariant distribution at the expense of technical requirements and mixing properties that might limit applicability in practice \citep{Maclaurin2014}. The alternative approach is to  run MCMC on small shards of the data and then construct a 'Consensus Monte Carlo' estimator \citep{Scott2013}\todo{lot more work on distributed MCMC other than consensus monte carlo (see references in Balajis NIPS paper). i think david dunson's work has some theory, would be good to check}.  At present the Consensus Monte Carlo algorithm lacks any theoretical guarantees.

In this contribution, we propose a different view on the problem. We construct a scheme that directly estimates posterior expectations with neither simulation from the full posterior nor construction of alternative approximate transition kernels -- \emph{without} introducing bias. While variance of these estimators is naturally increased, we will show that this increase is \emph{bounded by construction} and straightforward to control. This in particular holds for the Big Data context.
We arrive at the following desiderata for a useful methodology in unbiased Bayesian inference: (i) computational complexity that is sub-linear in the number of observations, (ii) controllable and bounded variance, and (iii) no problems with transition kernel design.

The presented framework is very general in the sense that it is neither restricted to a particular form of the underlying Bayesian model (such as factorising likelihoods), nor does it rely on a particular inference technique used from within. Any free parameters are easy to tune. Furthermore, it is competitive in practice: experimental examples show that we are able to accurately and confidently estimate posterior expectations, faster than with simulation methods.
In addition, and without exploiting any domain specific structure, we are able to outperform state-of-the-art results obtained from (approximate) stochastic variational inference methods on large-scale real-world data \citep{Hensman2013}.

By no means do we aim to \emph{replace} existing simulation (or any other) techniques for Bayesian inference. If the goal is to obtain a representation of the \emph{full} posterior density, simulation remains the method of choice.  Our contribution rather provides a different perspective on problems where a given Bayesian expectation lies at the core of the performed statistical analysis.

\paragraph{The setting} we consider is as follows. Given data ${\cal D} = \{x_1, \dots, x_N\}$, a statistical model with parameters $\theta\in \Theta\subseteq \mathbb R^p$, and likelihood $p(x_1, \dots, x_N|\theta)$, we wish to obtain an unbiased estimator of the expectation
\begin{align}
\mathbb E_{\pi_{N}}\{\varphi(\theta)\}
\label{eqn:estimation_problem}
\end{align}
 of a given functional $\varphi:\Theta\to \mathbb R$ over the full posterior $\pi_N=p(\theta|x_1, \dots, x_N)\propto p(\theta)p(x_1,\dots,x_N|\theta)$\todo{Clearly state assumptions}. While we focus on the real-valued $\varphi$ here, multivariate extensions are possible. A typical way to approach this problem is to generate samples $\left\{\theta_i\right\}_{i=1}^M$ from $\pi_N$ using MCMC, and then compute the empirical expectation $\hat{\mathbb E}_{\pi_{N}}\{\varphi(\theta)\}=\frac{1}{M}\sum_{i=1}^M \varphi(\theta_i)$. Note that the goal here is \emph{not} to obtain samples from the full posterior measure $\pi_N$ -- the focus is rather on the estimation of particular expectations. For example, in a regression setting, we might be interested in predictive posterior means and variances. This is the ubiquitous end goal in many situations in which Bayesian inference is employed. Therefore, we challenge the paradigm of solely working towards posterior simulation for such estimation problems, and propose a complementary methodology.

A subtlety when dealing with MCMC based estimates is that $\hat{\mathbb E}_{\pi}\{\varphi(\theta)\}$ for any posterior $\pi$ is only asymptotically correct. Therefore, any finite time MCMC algorithm produces estimates that contain a systematic bias, which  is subsequently made arbitrarily small via careful choice of simulation parameters. Parts of our methodology are based on such estimators. For the sake of simplicity, we will here assume access to the asymptotic limit in the same sense as finite time MCMC estimates are treated as 'correct'. We hint at a way to address this issue as an outlook. 

Moreover, our approach is not restricted to MCMC, but easily applies to situations where expectations over $\pi_N$ are available in closed form but require prohibitive amounts of computation. Such cases can for example be found in large-scale spatial statistics \citep{Lyne2013}, or Gaussian Process models deployed to Big Data regimes \citep{RasmussenCarlEdwardWilliams2006, Hensman2013}.

Our work builds on several breakthroughs in related areas, such as unbiased estimation for stochastic differential equations \citep{RheeGlynn2013} and for Markov chain equilibrium expectations \citep{GlynnRhee2014}. These developments demonstrate the overarching principle that estimation is often an easier problem than simulation -- a dictum we adopt and apply here in the context of Bayesian inference.

\paragraph{Paper outline}
We begin by summarising previous sub-sampling based simulation approaches in Section \ref{sec:previous_work}. In Section \ref{sec:PartialPosts}, we construct unbiased estimators for Bayesian expectations from paths of partial posterior distributions. Section \ref{sec:experiments} contains a number of experimental illustrations where we demonstrate that our estimator is in particular useful in the Big Data regime.  In Section \ref{sec:extensions}, we point out a number of extensions and conduct experiments that showcase the generality of the developed framework. We close with a discussion of shortcomings and point out future work in Section \ref{sec:discussion}.

\section{Previous work}
\label{sec:previous_work}
A practical difficulty in dealing with the full posterior $\pi_N=p(\theta|x_1, \dots, x_N)\propto p(\theta)p(x_1,\dots,x_N|\theta)$ is that $N$ is often large. This renders the computation of a likelihood $p(x_1, \dots, x_N|\theta)$ extremely expensive --  if not impossible. This, for example, limits feasibility of MCMC algorithms to simulate from $\pi_N$ as they require access to $p(x_1, \dots, x_N|\theta)$ in \emph{every iteration}.

\paragraph{Biased MCMC}
The infeasibility of exact likelihood computation has been the main focus of \citep{Welling2011,Korattikara2014,ChenFoxGuestrin2014,Bardenet2014} where this issue is circumvented by approximations to the transition kernel in MCMC. This is done using, e.g. stochastic gradient Langevin \citep{Welling2011} or Hamiltonian \citep{ChenFoxGuestrin2014} approaches, or by using a noisy acceptance ratio and employing a statistical test \citep{Korattikara2014} or concentration bounds \citep{Bardenet2014}. 
The well-known issue with this vein of work is that the approximate finite step-size diffusions, defined by {\em mini-batches} of data, are no longer corrected for induced bias. Consequently convergence to the correct posterior (and indeed any) measure \emph{is no longer guaranteed}. The practical effect of these approaches makes them difficult to tune and to obtain a well-mixing chain. Furthermore, artefacts of these methodologies, such as using parametric hypothesis testing  \citep{Korattikara2014}, might even lead to over-confident accept/reject decisions in the Markov transition kernel. The latter was illustrated in \citet{Bardenet2014}, who also substantially improve on these constructions by providing total variation bounds to assess the quality of the approximation.

\paragraph{Noisy MCMC}
Recently, \citet{Alquier2014} provided quantification of the approximation quality of many 'noisy' MCMC Algorithms, including the ones by \citet{Welling2011,Korattikara2014,Bardenet2014}. These results are an important step towards understanding the extent of the bias induced by employing approximate transition kernels. However, in practice their results require uniform or geometric ergodicity of the original Markov Chain to explicitly quantify the approximation error \citep[Theorem 2.1]{Alquier2014} or just guarantee convergence \citep[Theorem 2.2]{Alquier2014}, respectively. The first condition is too strong for most problems in practice while the second one does not give important details on how and when the approximate chain converges. Our work can be seen as an orthogonal approach, as we avoid simulation from the full posterior and rather directly attack the underlying estimation of expectations of interest, i.e. \eqref{eqn:estimation_problem}.

\paragraph{Firefly MCMC}
In contrast to these approximate, biased sampling methods, Firefly MCMC \citep{Maclaurin2014} introduces an exact construction that neither introduces bias nor requires computation of a full likelihood. It is an elegant way of exploiting additional auxiliary variable structure. In this regime of computationally intractable likelihoods due to data size\footnote{Data size as  opposed to computationally intractable likelihoods due to an inherently intractable functional defining the likelihood.}, it is seen as the only approach that can ensure coherence of subsequent inference through the simulation from the asymptotically correct posterior. One complication with Firefly MCMC is that it requires availability of appropriate {\em easily computable and tight} lower bounds on the likelihood function, tuning of which requires at least one sweep through \emph{all} data. Of course, the models for which such bounds can be obtained are often precisely those relatively simple models where the need for full and exact Bayesian inference over variational or other approximations might be questionable\todo{Add non-trivial example}. While investigating the formal construction of such bounds in more general model classes is a promising way forward, the generality and applicability of Firefly MCMC is clearly limited. Moreover, while Firefly MCMC can significantly reduce the number of likelihood evaluations at each iteration of MCMC, the complexity of the scheme is linear in the number of observations, as resampling of $q\cdot N$ auxiliary variables is required at each iteration, for a given fraction of the available data $ q\in (0,1)$. There is a limit as to how small the parameter $q$ can be chosen: mixing time \emph{cannot be smaller} than $1/q$. This means that the reduced number of likelihood evaluations at each iteration of MCMC comes at the cost of requiring to run the chains by a factor of $1/q$ longer.


 In contrast to all but one of the previous sub-sampling schemes considered, the estimators that we propose are provably unbiased and also have a sub-linear average complexity in the number of observations. Unlike the only unbiased competitor, Firefly MCMC, our approach does not require a lower bound on the likelihood and even extends to several other situations: where posterior expectations are available in closed form but computationally infeasible, where likelihoods need not factorise across observations, and where likelihoods might themselves be unavailable in an analytic form, as for example in the context of pseudo-marginal MCMC \citep{Andrieu2009}.

\section{Partial posterior path estimators}
\label{sec:PartialPosts}
 In this section, we present a different approach to coherent Bayesian inference in the Big Data regime which exploits the paths of induced partial posterior distributions through the debiasing device developed in \citet{RheeGlynn2013, GlynnRhee2014}.
A similar approach was very recently taken by \citet{Agapiou2014}, who exploit \citeauthor{RheeGlynn2013}'s work for unbiased posterior estimation of expectations over intractable infinite-dimensional models which can be parametrised in terms of a series expansion of basis functions. In contrast, our work directly attacks intractability that arises from large datasets. We see our contribution as a pragmatic complement to existing work on debiasing Monte Carlo estimates.

Our approach follows similar ideas as \citet{Chopin2002}, who presented a sequential Monte Carlo procedure for static target distributions by exploiting a sequence of partial posterior targets. Given the full posterior $\pi_N= p(\theta | x_1, \dots, x_N) \propto p(x_1, \dots, x_N|\theta)p(\theta)$, we define a subset $\mathcal{D}_t=\{x_i\}_{i \in \mathcal{I}_t}$  of size $n_t=|\mathcal{I}_t|$ of all data, where $\mathcal{I}_t \subseteq \{1,\dots,N\}$ is a (possibly random) index set\todo{this footnote doesn't make sense - haven't introduced $T$ yet}, with sizes  $0<n_1<n_2<\dots n_L=N$. The partial posterior corresponding to $\mathcal{I}_t$ is then $\pi_t = p(\theta | \mathcal{D}_t) \propto p(\mathcal{D}_t|\theta)p(\theta)$\footnote{Note that while $L$ is the number of subsets, they are indexed with the subscript $t=1 \dots, T$ for some variable $T\leq L$ that will be introduced later.}. Paths of partial posterior measures can be constructed by starting from the prior $\pi_0(\theta)=p(\theta)$ and increasing the size of the batches $\mathcal{D}_t$ until exhausting the whole set of observations, i.e.
\begin{align*}
\pi_0(\theta)\rightarrow \pi_1(\theta)\rightarrow \pi_2(\theta)\rightarrow \dots \rightarrow \pi_N(\theta),
\end{align*}
where $\pi_N(\theta)=p(\theta|x_1,\dots,x_N)$ is the full posterior.


For simplicity of exposition, we here consider the case of a geometric increase in batch sizes. More precisely, we set $n_1=a,\dots,n_t=2^{t-1}a,\dots,n_{L}=2^{L-1}a=N$, where $L=\log_2(N/a)+1$ is the number of possible batch sizes, $a$ is the smallest batch size considered. We assume that $\log_2(N/a)$ is an integer. Note that common ratios other than 2 are possible, and are used in the experiments.

The next section presents the debiasing device, which is a key component in transforming estimates of expectations over multiple partial posteriors ${\mathbb E}_{\pi_t}\{\varphi({\theta})\}$  into an unbiased estimator of the expectation over the full posterior ${\mathbb E}_{\pi_N}\{\varphi({\theta})\}$.

\subsection{Debiasing Lemma \& algorithm}
\label{sec:Lemma_and_algo}
The debiasing Lemma provides a way to construct an unbiased estimator for the limit of a converging a sequence -- without evaluating all elements. Here, we transform a sequence of asymptotically correct, but biased estimators into an unbiased estimator. In different contexts, the result was originally discussed independently by \citet{McLeish2011} and \citet{RheeGlynn2012}. It was shown in the present form in \citet[Theorem 1]{RheeGlynn2013}; see also \citet[Theorem 1.1]{JacobThiery2013}.

\begin{lem}[Telescoping Estimator]
\label{eq:telesoping_estimator}
Let $\phi$ and $\left\{ \phi_{t}\right\} _{t=1}^{\infty}$ be real-valued random variables\footnote{\citet{Agapiou2014} very recently developed a Hilbert space version of the Lemma.}, and let $T$ be an integer-valued random variable independent of $\phi$ and of $\left\{ \phi_{t}\right\}_{t=1}^{\infty}$
with $\mathbb{P}\left[T\geq t\right]>0$ $\forall t\in\mathbb N$. With the convention $\phi_{0}=0$, assume that 
\begin{equation}
\sum_{t=1}^{\infty}\frac{\mathbb{E}\left\{\left|\phi_{t-1}-\phi\right|^{2}\right\}}{\mathbb{P}\left[T\geq t\right]}<\infty.\label{eq:assumption_convergence}
\end{equation}
Then, 
\[
\phi_{T}^{*}  =  \sum_{t=1}^{T}\frac{\phi_{t}-\phi_{t-1}}{\mathbb{P}\left[T\geq t\right]}
\]
is an unbiased estimator of $\mathbb{E}\{\phi\}$, i.e. $\mathbb{E}\{\phi_{T}^{*}\}=\mathbb{E}\{\phi\}$. Moreover,
\[
\mathbb E \left\{\left(\phi_{T}^{*}\right)^2\right\}=\sum_{t=1}^{\infty}\frac{\mathbb{E}\left\{\left| \phi_{t-1}-\phi\right|^{2}\right\}-\mathbb{E}\left\{\left| \phi_{t}-\phi\right|^{2}\right\}}{\mathbb{P}\left[T\geq t\right]}.
\]
\end{lem}

\todo{HS: We should have a proof sketch of the unbiasedness here, or at least reference the Appendix}


%

\paragraph{Finite variance for unbiased Bayesian expectations}
The above Lemma \ref{eq:telesoping_estimator} is directly applicable in our context. We set $\phi_t=\hat{\mathbb E}_{\pi_t} \{\varphi\left(\theta\right)\}$, for $t<L$ where the empirical expectation $\hat{\mathbb E}_{\pi_t}$ is computed by, e.g. MCMC\footnote{We realise that empirical expectations computed with MCMC are technically biased and will comment on this in the next Section.} on the partial posterior $\pi_t$, and $\phi_t=\phi=\hat{\mathbb E}_{\pi_{N}} \left\{\varphi\left(\theta\right)\right\}$ for $t\geq L$. In the finite data regime, the conditions of the above Lemmas are trivially satisfied since we set $\phi_t=\phi$ almost surely for $t\geq L$. The variance of estimators is thus \emph{finite by construction} and the truncation variable $T$ needs only to be supported on $1,\dots,L$, and all infinite sums can be replaced with sums up to $L$. However, variance might still increase increase with $N$ without a bound. Therefore, in order to ensure stability of the estimators in the Big Data regime, we here require an analogous condition to \eqref{eq:assumption_convergence} that will guarantee that the variance remains constant, i.e. that as $N\to\infty$, \todo{Check expectation arguments}
\begin{equation}
\sum_{t=1}^{L}\left(\frac{\mathbb{E}\left\{\left|\hat{\mathbb E}_{\pi_{t-1}}\{\varphi(\theta)\}-\hat{\mathbb E}_{\pi_{N}}\{\varphi(\theta)\}\right|^{2}\right\}}{\mathbb{P}\left[T\geq t\right]}\right)=\mathcal{O}(1).
\label{eq:assumption_bayes}
\end{equation}
Intuitively, we require that the tail of the stochastic truncation variable matches the rate of convergence of partial posterior expectations. See the next Section and Appendix \ref{sec:Appendix_complexity_variance} for details on a simple setup where this condition holds, along with a way to tune the truncation distribution $\mathbb{P}[T\geq t]$.

Note also that in the same fashion as in \citet{RheeGlynn2013}, one can replicate the random truncation procedure $R$ times and thus reduce variance. More precisely, $\frac{1}{R}\sum_{r=1}^R \phi_{T_r}^{*}$ is an unbiased estimator of $\mathbb E_{\pi_{N}}\{\varphi(\theta)\}$ and its variance scales as $1/R$. Here, $\{T_r\}_{r=1}^R$ are independent copies of $T$ and each $\phi_{T_r}^{*}$ is computed on a different partial posterior path.  This implies that the scheme can be repeated until a desired error tolerance is attained. The latter can be estimated from the empirical variance of the  $\phi_{T_r}^*$.

\begin{figure}
\centering
\includegraphics{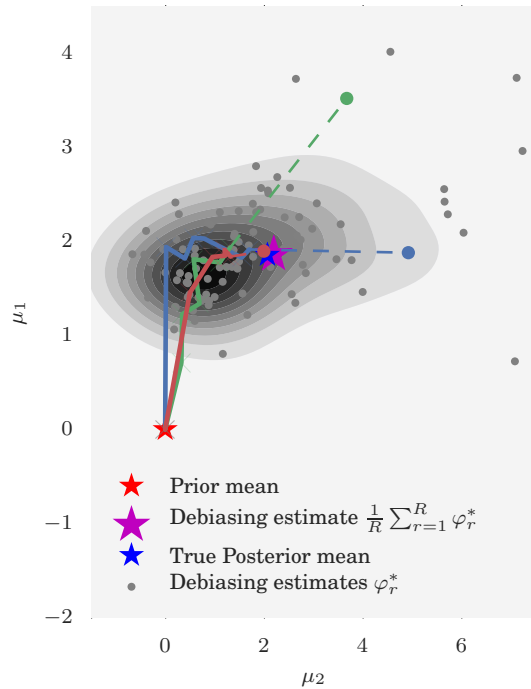}
\caption{Illustration of Algorithm \ref{alg:debiasing} for the posterior mean of a 2D Gaussian with unknown mean $\mu$ and fixed covariance $\Sigma$. Data is $\mathcal{D}=\{\mathbf{x}_i\sim\mathcal{N}(\mathbf{x}_i|\mu=\mathbf{2},\Sigma)\}_{i=1}^{100}$ with $\Sigma=[(-1,3)^\top,(3,1)^\top]$, prior is $p(\mathbf{\mu})=\mathcal{N}(\mathbf{\mu}|\mathbf{0}, I)$. We aim to compute the posterior mean $\int \mu p(\mu|\mathcal{D})d\mu$.
Debiasing computes multiple posterior paths (coloured solid lines), which are randomly truncated (solid line stops), and then plugged into the debiasing estimator in \eqref{eq:telesoping_estimator} to estimate the posterior mean of $\mu_1$ and $\mu_2$ (coloured round dots, dotted lines connect path end-point to estimate). The procedure is averaged $R=1000$ times (gray dots), after which the empirical mean matches the full posterior mean. A kernel density estimate of the gray dots is shown in the background. }
\label{fig:posterior_path}
\end{figure}

Algorithm \ref{alg:debiasing} summarises our approach, and Figure \ref{fig:posterior_path} illustrates both construction of the $\phi_{T_r}^*$ from partial posterior paths, and their distribution.

\begin{algorithm}
\caption{Debiasing posterior expectations}
\label{alg:debiasing}
\textbf{Input:} Discrete distribution $\Lambda$ over $1\dots,L$, corresponding to batch indices\\
For $r=1,\dots,R$ (number of replications), or\\while not achieved desired error tolerance
 \begin{itemize}
  \item Sample truncation variable $T_r\sim \Lambda$
  \item For $t=1,\dots, T_r$
  \begin{itemize}
   \item Compute $\phi_t=\hat{\mathbb E}_{\pi_t} \{\varphi\left(\theta\right)\}$ (expectation over the partial posterior on batch $\mathcal{D}_t$ of size $n_t$), e.g. by running MCMC on $\pi_t$
  \end{itemize}
  \item Compute the debiased estimate $\phi_{r}^{*} = \sum_{t=1}^{T_r}\frac{\phi_{t}-\phi_{t-1}}{\mathbb{P}\left[T_r\geq t\right]}$
 \end{itemize}
\textbf{Return:} the average of debiased estimates $\phi^*=\frac{1}{R}\sum_{r=1}^R \phi_{r}^{*}$.
\end{algorithm}
\todo{Peter D: Add a Figure to Figure \ref{fig:posterior_path} that explains a \emph{single} debiasing estimator, i.e. how the weighting corrects for the stopping}

In summary, the key properties of the described methodology is that full posterior expectations over $\pi_N$ can be estimated, with \emph{no bias} introduced and with a \emph{bounded} increase in variance. This is achieved by using sub-samples of the available data -- at a sub-linear average computational cost as we will see next.

\subsection{Practical considerations}
\label{subsec:practical_considerations}
We now list several properties, implications, and key advantages of our scheme.

\todo{Balaji suggest to think about the subsampling the data: Do the batches overlap or not? What does this mean for the estimators? He told us to look at \citep{Kleiner2014}.}

\todo{Balaji:  the tradeoff between $L$ and $R$ wasn't clear to me. does the theory suggest a good default value? (e.g. iirc, the bag of little bootstraps \citep{Kleiner2014} suggests something like each bag should use $N^0.6$). HS: We should ellaborate on that, I think he got confused there.}

\paragraph{Computational costs and variance}
Let us denote by $\tau$ the time required to generate a single debiasing estimator $\phi_{T}^{*}$. Since computing $\phi_{T}^{*}$ requires running $T$ MCMC chains, on batch sizes $n_1,\dots,n_T$, $\tau$ would scale linearly with the overall number of likelihood evaluations, resulting in the average time complexity of $\mathbb E\{\tau\}=\mathcal{O}(\mathbb E_{T}\{n_1+...+n_T\})$. If the batch-size increase is geometric, i.e. $n_1=a,\dots,n_t=2^{t-1}a,\dots,n_{L}=2^{L-1}a=N$, the cost becomes $\mathcal{O}\left(a\mathbb E_{T}\left\{2^T\right\}\right)$.
By matching this with truncation probabilities $\Lambda_t=\mathbb P\left[T=t\right]\propto 2^{-\alpha t}$, for $0<\alpha<1$, we obtain an average cost of $\mathcal{O}\left(a(N/a)^{1-\alpha}\right)$, which is \emph{sub-linear in $N$}, see also Appendix \ref{sec:Appendix_complexity_variance}. This cost reflects the amount of computation when only a single core is available, and the trivial parallelisation of the scheme allows further savings, as described below.

The variance of the debiasing estimator depends on the rate of convergence of the partial posterior expectations. In order to ensure that te variance stays bounded as $N$ increases, assume that there exist a constant $c$ and $\beta>0$, such that for large enough $N$ and $\forall t\leq L$:
\begin{equation*}
 \mathbb{E}\left\{\left|\hat{\mathbb E}_{\pi_t}\{\varphi(\theta)\}-\hat{\mathbb E}_{\pi_{N}}\{\varphi(\theta)\}\right|^{2}\right\}\leq \frac{c}{n^\beta_{t}}.
\end{equation*}
 From here, as shown in Appendix \ref{sec:Appendix_complexity_variance}, \eqref{eq:assumption_bayes} is satisfied and therefore variance remains bounded as long as $\alpha<\beta$. Thus, fast convergence of partial posterior expectations, e.g., $\beta$ close to $1$, can result in significant speed-ups of the scheme. We give examples of empirical fits for $\beta$ in Appendix \ref{sec:Appendix_complexity_variance}.

\begin{figure}
\centering
\includegraphics{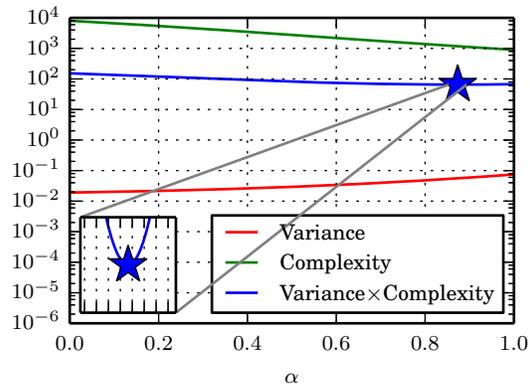}
\caption{Complexity-Variance tradeoff as a function in $\alpha$ for $N=10000$. Optimising these quantities for a minimum batch size of $a=128$ gives a best $\alpha=0.87$. The zoomed-in area shows a zoom-in around the minimum (with a different scaling on both axes) and reveals that the minimum is well defined.}
\label{fig:work_variance_product}
\end{figure}

\paragraph{Tuning truncation probabilities}
We now describe how to tune free parameters of the scheme. Following \citet{GlynnWhitt1992}, if both the average time complexity $\mathbb E \{\tau\}$ and the variance $\text{Var}\left\{\phi_{T}^{*}\right\}$ are finite, a central limit theorem holds in the limit where computational budget $\kappa\to\infty$. Namely, for a given computational budget $\kappa$, if we denote by $R_\kappa$ the number of debiasing replications that can be generated in $\kappa$ time, $R_\kappa=\max\left\{R\geq 0 : \sum_{r=1}^{R}\tau_r\leq \kappa\right\}$ and by $\phi_{(\kappa)}^*=\frac{1}{R_\kappa}\sum_{r=1}^{R_\kappa} \phi_{r}^{*}$ the resulting average of debiasing replications, then
\begin{equation}\label{eq:clt_work_variance}
 \sqrt{\kappa}\left(\phi_{(\kappa)}^*-\mathbb E\{\varphi(\theta)\}\right)\rightsquigarrow \mathcal N\left(0,\mathbb E \{\tau\} \text{Var}\left\{\phi_{T}^{*}\right\}\right). 
\end{equation}
Thus, the asymptotic efficiency of the debiasing estimator is characterised precisely by the work-variance product. Figure \ref{fig:work_variance_product} demonstrates how the distribution of the stochastic truncation variable (here in the parametric form $\mathbb P\left[T=t\right]\propto 2^{-\alpha t}$) can be optimised in order to yield minimal asymptotic variance of the estimators, see also Appendix \ref{sec:Appendix_complexity_variance}.

\paragraph{MCMC and finite time bias}
Any empirical expectation computed from finite MCMC algorithms is only correct in the asymptotic limit. Consequently, setting $\phi_t=\frac{1}{M}\sum_{i=1}^M \varphi(\theta_i)$ for $\theta_i$ sampled from a finite time Markov chain is problematic, as unbiasedness of the overall approach technically corrupted. However, the same is true for \emph{any} MCMC estimate. In practice, careful tuning of simulation parameters such as burn-in, thinning, and running multiple chains \citep{gelman1992inference}, is successfully used to reduce finite time biases to a neglectable level. We adopt this mindset here for the sake of practicality and simplicity of presentation. A way to address the issue could be to apply debiasing to the Monte Carlo estimate itself. In Lemma \ref{eq:telesoping_estimator}, set $\phi_{t,\ell}=\frac{1}{M}\sum_{i=1}^M \varphi(\theta_{i,\ell})$ for $\theta_{i,\ell}$ drawn from an approximate (not converged) Markov chain after $\ell$ iterations, as opposed to be drawn from the asymptotically invariant distribution. This gives a sequence $\{\phi_{t,\ell} \}_{\ell=1}^{\infty}$. Via applying the debiasing Lemma, an unbiased estimator  $\hat{\mathbb {E}}_{\pi_t} \{\varphi\left(\theta\right)\}$ can be constructed for any partial posterior $\pi_t$'s expectation. In a second stage, these debiased partial posterior expectation estimates can be used to estimate the full posterior expectation -- now fully unbiased. Unfortunately, as $\{\phi_{t,\ell} \}_{\ell=1}^{\infty}$ is an infinite sequence, the variance expression in Lemma \ref{eq:telesoping_estimator} is not trivially guaranteed to be finite anymore. See \citet{GlynnRhee2014, McLeish2011, Agapiou2014} for an explicit and in-depth treatment of unbiased Monte Carlo estimation.

\paragraph{MCMC and mixing time} If a Markov chain is, in line with above considerations, used for computing partial posterior expectations $\mathbb{E}_{\pi_t}\{\varphi\left(\theta\right)\}$, it need not be induced by any form of approximation, noise injection, or state-space augmentation of the transition kernel. As a result,  the notorious difficulties of ensuring acceptable mixing and problems of stickiness are conveniently side-stepped -- which is in sharp contrast to all existing approaches.
Furthermore, while the latter are limited to particular MCMC schemes \citep[random walk]{Bardenet2014}, and \citep[Langevin]{Welling2011}, \emph{any} MCMC procedure can be employed in our construction. This allows us to harvest decades of mathematical and engineering effort that went into both methodology and software packages (e.g. Stan  \citep{stan2014}.
Mixing time when using MCMC to estimate partial posterior's expectations is not compromised by our approach, in contrast to for example Firefly MCMC, whose mixing time gets worse as the mini-batch size decreases. MCMC chains over partial posteriors do not suffer from such problems. Indeed they are in practice often easier to handle due to their similarity to the (usually simply structured) prior distribution. 
\todo{Peter D: He did not get that we talk about mixing problems come from the augmented transition kernels, so ellaborate on that}

\todo{Balaji: - mixing time: you talk about mixing time about firefly MC ... i'd guess mixing would affect posterior expectations of functionals less than the scenario where you want to generate samples from posterior. maybe it might be better to discuss bias variance of firefly MC than mixing time.}


\paragraph{Parallelisation} As computation required for each partial posterior $\pi_t$ in a single path can be performed independently, the method embarrassingly parallelises and expectations computed in parallel only need to be combined in a telescoping sum. The same holds true for replications of the scheme: since the computational cost within each truncated posterior path is dictated by the largest batch size that needs to be processed in parallel, the required wall-time in the case of geometric batch-size increases is roughly halved. Therefore, with sufficient computational resources, the potential speed-up factor through parallelisation is $2R$, where $R$ in practice is usually in the 100s to 1000s as we will see in the experiments.\todo{Reference Appendix for derivation}

\section{Experiments}
\label{sec:experiments}
In this section, we illustrate the utility of the debiasing approach and compare it against other unbiased approaches: full posterior sampling  and Firefly MCMC. In particular, we show that for large-scale datasets, debiasing can accurately and confidently estimate posterior expectations \emph{before full MCMC and Firefly have produced a single estimate}. 

It is clear that running an MCMC chain on the full posterior, for any statistic, produces more accurate estimates than the debiasing approach, which by construction has an additional intrinsic source of variance. This means that if it is possible to produce even only a \emph{single} MCMC sample (after burn-in), the resulting posterior expectation can be estimated with less expected error. It is therefore not instructive to compare approaches in that region.
\todo{Why is it clear? Reference?}

\paragraph{On comparing to Firefly MCMC}
For a fair comparison of our method to Firefly MCMC, we give an estimate for the number of likelihood evaluation necessary for Firefly MCMC to produce the first sample -- for which there are two notable obstacles. The first is computing a maximum a posteriori (MAP) estimate to initialise a lower bound on the likelihood: \citet{Maclaurin2014} reported Firefly's performance  to be inferior to standard MCMC otherwise. For large datasets, MAP estimates are challenging as a standard gradient based optimisation scheme such as BFGS needs multiple evaluations of the full likelihood. 
For example, Stan's BFGS implementation on a commonly used benchmark dataset, \texttt{a9a} \citep{Welling2011, Lin2008}, takes around 40 iterations to reach a reasonable convergence level.
While this issue can be somewhat sidestepped via using stochastic gradient descent. However, given a MAP estimate, a one-off cost of $\mathcal{O}(N D)$, i.e. computing sufficient statistics of \emph{all} data \citep{Maclaurin2014}, \emph{cannot} be avoided. This is challenging for extremely large datasets.
Furthermore, Firefly is based on binary indicator variables that determine whether a point in a factorised likelihood is used for an MCMC update (bright) or not (dark). The second obstacle comes from Firefly's parameter that is the probability of brightening a dark point. First, at least $q_{\text{d}\rightarrow\text{b}}N$ points need to be evaluated in each iterations, which is \emph{linear} in $N$. Second, mixing time suffers at least by a factor of $1/q_{\text{d}\rightarrow\text{b}}$, which means that burn-in time and number of MCMC iterations need to be multiplied by that factor to compare to full MCMC in a fair way. Together, this means that Firefly MCMC roughly needs the same number of likelihood evaluations as full MCMC before it produces the first sample -- implying that our comparisons to the full MCMC directly apply to Firefly MCMC as well.

We now provide a number of examples, where we analyse convergence of the debiasing estimator up to the number of likelihood evaluations necessary to produce a single sample. All estimates are given as function of the number of likelihood evaluations needed to compute them (including burn-in). Note that, in favour of competing methods, we do not take parallelisation into account, which (given appropriate hardware) would increase the effective number of likelihood evaluations per unit time by a factor of $2R$.


\subsection{Synthetic log-Gaussian}
We first consider a toy model from \citet{Bardenet2014}, but with more data\footnote{Attempting to resist the commonly followed temptation of applying large-scale methodology to only medium sized datasets. 

}: rather than the original $10^5$, we generate $2 ^ {26} \approx 10^8$ data from a log-Gaussian $\log\mathcal{N}(\mu,\sigma^2)$, where $\mu=0$ and $\sigma^2=2$. Using flat priors, we sample from the joint posterior over $(\mu,\sigma)$ and aim to estimate the marginal posterior mean of $\sigma$.
This posterior has extremely wide tails, which causes problems for MCMC methods based on approximate transition kernels. In particular, \citeauthor{Bardenet2014} illustrate that even when using an appropriate setting of the mini-batch size, \citeauthor{Korattikara2014}'s scheme (confidently) gives completely wrong results. \citeauthor{Bardenet2014}'s sampler is able to recover the model's standard deviation but this comes at the cost of using almost  \emph{all} available data in \emph{every} MCMC iteration.

\begin{figure}
\centering
\includegraphics{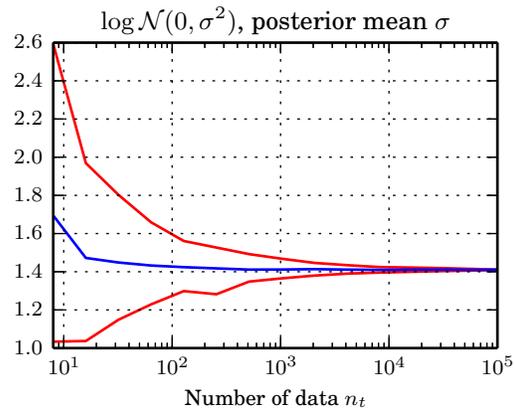}
\caption{Convergence of partial posteriors' mean of $\sigma$ as a function of sub-sample size for \citeauthor{Bardenet2014}'s log-Gaussian model. We randomly sub-sample $\eta_t$ data from the dataset and run MCMC to estimate $\sigma$, whose true value is $\sigma=\sqrt{2}$. Error bars are 95\% over 150 trials.}
\label{fig:bardenet_convergence}
\end{figure}

This happens despite the fact that such simple posterior expectations converge rapidly. Figure \ref{fig:bardenet_convergence} shows convergence of the partial posterior mean of $\sigma$ as a function of mini-batch size $\eta_t$. Even though all such estimates are biased, the plot reveals that using multiple MCMC chains on subsets of constant size and averaging gives a small estimation error quickly. This raises the question whether manipulating the Markov transition kernel  is the best way of addressing such problems.

\begin{figure}
\centering
\includegraphics{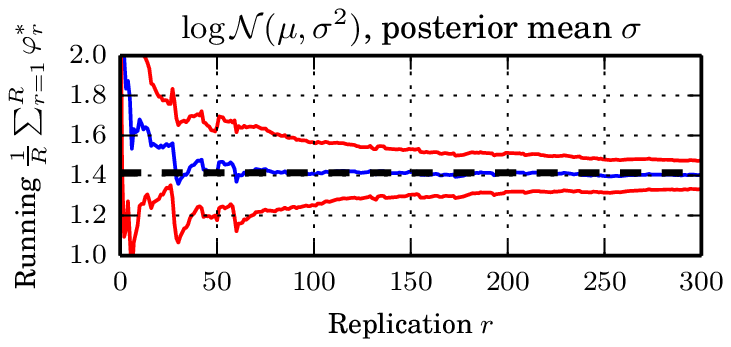}
\includegraphics{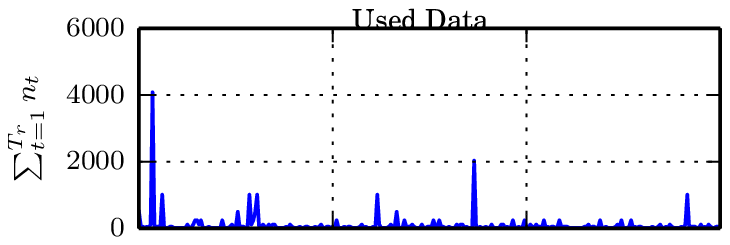}
\caption{Convergence and used data for debiased mean of posterior mean of $\sigma$ for \citeauthor{Bardenet2014}'s Log-Gaussian model. \textbf{Top:} Debiasing estimates converge to true posterior statistic $\sigma=\sqrt{2}$ quickly. \textbf{Bottom:} Small per-replication and cumulative data usage. }
\label{fig:debiasing_bardenet_log_gaussian}
\end{figure}

Debiasing, in contrast, is a way of exploiting rapid convergence of posterior expectations, while remaining unbiased -- as demonstrated in the upper part of Figure \ref{fig:debiasing_bardenet_log_gaussian},  where we show that we can recover the true model parameter confidently and quickly. We run a number of debiasing replications with a minimum batch size of $a=8$, and geometric truncation probability $\alpha$ close to $1$. Each of the partial posterior expectations are computed via MCMC\footnote{We use the NUTS sampler \citep{stan2014}, assuming a constant number of leap-frog iterations in HMC.} with $500$ iterations after a burn-in of $100$ iterations. We count the number of likelihood evaluations for each partial posterior, taking into account the $600$ MCMC iterations. We run full MCMC with the same number of iterations and burn-in (note that this is in favour of full MCMC as partial posterior distributions should be explored in fewer iterations). We count $N=2^{26}$ likelihood evaluations per MCMC iteration, with an offset of $100 N$ burn-in iterations.

Remarkably, as Figure \ref{fig:debiasing_bardenet_log_gaussian} indicates, the \emph{largest} partial posterior size was only $\max_{r} \left\{n_{T_r}\right\}=2048$, leading to a maximum single replication cost of $\sum_{t=1}^{\max_r \{T_r\}}n_t= 4088$ as depicted in Figure \ref{fig:debiasing_bardenet_log_gaussian}, and a median $n_{T_r}$ of only $16$. After $R=300$ replications, the number of data touched in total is $\sum_{r=1}^{R} \sum_{t=1}^{T_r} n_t=27,264$. Taking into account the $600$ MCMC iterations to estimate each partial posterior expectation, this sums up to $16,358,400$ likelihood evaluations, which is \emph{less than a quarter} of a single full MCMC burn-in iteration ($N=2^{26}$), and less than $1/(4\cdot 500)\approx 0.0005$ times the number of likelihood evaluations required to complete the burn-in of full MCMC.

Fast convergence of posterior expectations as in Figure \ref{fig:bardenet_convergence} is not an artefact of the above model. As we demonstrate in the next experiment, such situations arise in more involved inference tasks as well.

\subsection{Large-scale logistic regression}
\label{sec:logistic_toy_experiment}
We now apply our methodology to a large-scale Bayesian logistic regression problem on $N=10^8$ data. As full posterior simulation is infeasible for models of such size, we choose a synthetic dataset in order to quantify estimation error.

 We model binary labels of $N=10^8$ observations of $D=9$ features $x_i\in\mathbb{R}^{D}$ as $p(y,\beta)=\sigma(y_i \beta^\top x_i)p(\beta)$, where $\sigma$ is the logit function and $p(\beta)$ are independent Laplace priors with a scale of 1. The bias parameter is absorbed into $\beta_{10}$ and $x_{10}=1$. To generate data, we sample covariates $x_i \sim \mathcal{N}(x_i|0,D^{-2})$ and label them positively with probabilities $\sigma(x_i^\top \beta)$. True regression weights are set to $\beta_i=1$ for $i=1,\dots,D$.
 
\begin{figure}
\centering

\includegraphics{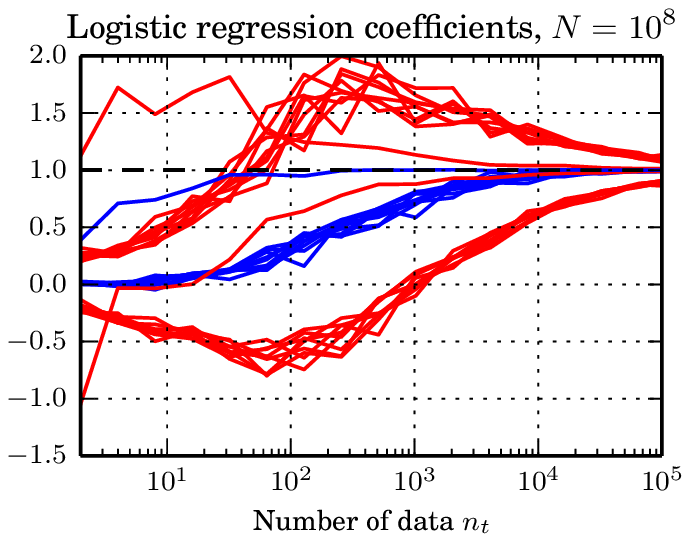}
\includegraphics{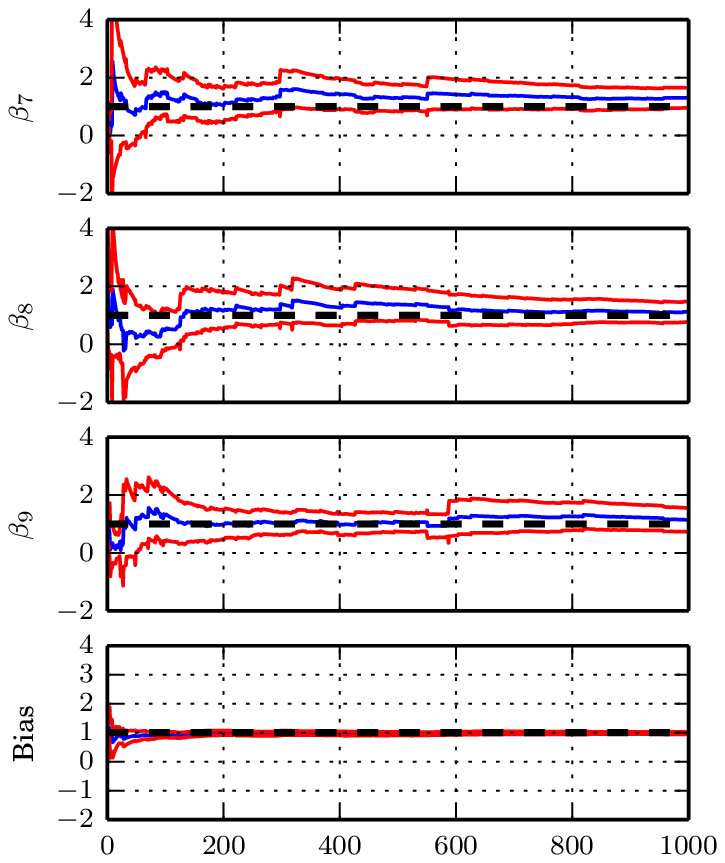}
\caption{Logistic regression on $N=10^8$ synthetic data. \textbf{Top:} Posterior mean convergence as a function of mini-batch size, for all 9 regression weights and the bias term (outlier in the plot). They converge from about 10000 data (averaged over 100 trials). Note how the bias term compensates for small weights in the low data regions of the plot. \textbf{Bottom:} Debiasing convergence as a function of the number of replications $R$, for the last 3 regression weights and bias term. Other regression weights behave similarly. 95\% error bars, dashed line indicates ground truth from the above plot. All $R=1000$ replications correspond to $9 N$ likelihood evaluations, not taking parallelisation into account.}
\label{fig:logistic_toy}
\end{figure}

 As in the previous example, simple posterior statistics such as mean regression weights, i.e. $\varphi(\beta)=\beta_i$ for $i=1,\dots,10$, converge quickly: Figure \ref{fig:logistic_toy} (top) reveals that the statistics do not significantly change when computed from randomly sub-sampled mini-batches larger than 10000, which is 4 order of magnitudes smaller than $N=10^8$.
  Note that it not possible to run even a single MCMC chain on the full dataset -- in contrast to our debiasing approach.

We apply our debiasing estimator, using a minimum batch size of only $a=100$, a geometric batch size increase of $2$, and run Stan's NUTS sampler for 500 iterations after a burn-in of 100 iterations. Figure \ref{fig:logistic_toy} (bottom) shows examples of the convergence of the debiasing estimator over $R=1000$ replications. Taking into account the 500+100 inherent MCMC iterations, the median data usage per replication is $256\cdot 600$ likelihood evaluations data points, the average is $722\cdot 600$. Summing over all $1000$ replications, debiasing takes $0.02N\cdot 600 \approx 9  N$ likelihood evaluations. This means that a full MCMC chain would not even have passed the $100$ iterations of burn-in, while debiasing already converged close to the ground truth posterior statistic. We stress that this comparison is extremely conservative: given appropriate computational resources, parallelisation allows for a speed-up of up to factor $2R=2000$. This means that we can reduce error bars by an additional large factor without increasing computation wall-time.

%
%
%
%

\section{Extensions}
\label{sec:extensions}
We now describe two extensions of our framework that illustrate its generality compared to other sub-sampling based approaches, and give experimental illustration.  This includes an experiment where we are able to outperform stochastic variational inference for Gaussian Processes on a large-scale real-world dataset.

\subsection{Likelihoods need not factorise}
The debiasing device for constructing the unbiased estimators of posterior expectations \emph{does not} require that the likelihood factorises, i.e. that
\begin{align*}
p(x_1,\dots,x_N|\theta)=\prod_{i=1}^N p(x_i|\theta).
\end{align*}
We simply require access to partial likelihoods $p(x_1,\dots,x_{n_l}|\theta)$ for a given batch size $n_l$.
To the best of our knowledge, this is in sharp contrast to all other methods available for MCMC in the Big Data regime, where the likelihood has to be computable and typically is assumed to factorise.
As such, debiasing over partial posterior paths can also be applied to cases where posterior distributions are available in closed form -- but only at a prohibitive amount of computational cost.

\paragraph{Approximate Gaussian Process regression}
is a typical example for a non-factorising likelihood. We focus on a simple 
case of predictive posterior in Gaussian Process (GP) regression,
\begin{align}
&&&\pi_N\big(&&y_*|x_*,y,X\quad\big):=p(y_*|x_*,y,X)\notag\\
&&=\text{ }&\mathcal{N}\Big(&&k_*^\top(K+\lambda I)^{-1}y,\notag\\
&&&&&k(x_*,x_*)- k_*^\top(K+\lambda I)^{-1}k_*\quad\Big),
\label{eq:gp_predictive_posterior_dual}
\end{align}
where $K$ is the covariance function evaluated at pairwise training covariates $X$, and $k_*=(k(x_1,x_*),\dots,k(x_N,x_*))^\top$, and observation noise variance $\lambda$ \citep[Section 2]{RasmussenCarlEdwardWilliams2006}. This requires the inversion of an $N\times N$ covariance matrix and therefore costs $\mathcal{O}(N^{3})$ computation.
Note that predictive mean and 
variance here can be computed exactly -- no MCMC simulation
is required. With the debiasing approach, it suffices to look at the expectations
of partial predictive posteriors $\pi_{j}(y_{*})$, which is again based on sub-sampling all available data $X$ and $y$. As each evaluation then requires $\mathcal{O}(n_{j}^{3})$
computation, the average computational costs are given by $\mathbb{E}_{T}\left(\sum_{j=1}^{T}n_{j}^{3}\right)=\mathcal{O}\left(N^{3-\alpha}\right)$, 
where we set $n_{j}=2^{j-1}a$, and $p_{t}\propto2^{-\alpha t}$ as
before.

The above, however, can still be infeasible in practice, and further savings can be obtained by 
applying the debiasing onto the primal form of \eqref{eq:gp_predictive_posterior_dual}, in combination with
an explicit finite rank representation of the kernel function $k(x,x')=\phi_{x}^\top\phi_{x}$,
with $\phi_{x}\in\mathbb{R}^{m}$, e.g. inducing variables \citep{QuinoneroCandela2005}, random
Fourier features \citep{Rahimi2007}, or Incomplete Cholesky \citep{Fine2001}. By
performing Bayesian linear regression on this explicit (approximate) feature space,
the posterior for a single test feature $\phi_*$ becomes
\begin{align}
\label{eqn:predictive_gpr}
\pi_{N}(y_{*}|\phi_*,\Phi,y) & = &  \mathcal{N}\Big(\phi_{*}^{T}\left(\Phi^{T}\Phi+\lambda I\right)^{-1}\Phi^{T}y,\nonumber\\
 &  & \qquad\phi_{*}^{T}\left(\Phi^{T}\Phi+\lambda I\right)^{-1}\phi_{*}\Big),
\end{align}
with feature matrix $\Phi=[\phi_1^\top,\dots,\phi_N^\top]^\top$. Evaluation now requires a reduced cost of $\mathcal{O}(m^{2}N)$. Having a cost linear in $N$ for obtaining each expectation gives a debiasing average computational cost of  $\mathcal{O}\left(m^{2}N^{1-\alpha}\right)$, which again is sub-linear in $N$ for a fixed feature space dimension $m$.
As before, each of the mini-batches can be processed in parallel.

\paragraph{Experimental illustration}
To illustrate the above idea, we generate $N=10^4$ toy data for a univariate non-linear regression problem in an approximate feature space given by \citeauthor{Rahimi2007}'s random Fourier features. Note that this exactly corresponds to a Bayesian linear regression with the mapped features. More specifically, we choose a Gaussian kernel with unit length scale, $k(x,x')=\exp\left(-\Vert x-x'\Vert^2\right)$, whose associated random feature space of dimension $m=100$ is given by the mapping
\begin{align*}
\sqrt{m}\phi_x = (\cos(w_1 x +b_1),\dots,\cos(w_m x +b_m))^\top,
\end{align*}
where $w_i \sim \mathcal{N}(0,1)$ and $b_i\sim\texttt{Uniform}(0,2\pi)$ are fixed and the covariates are randomly spread in $[0,10]$. We sample a set of training labels from the corresponding approximate GP prior, add observation noise, and resample the feature space basis via $w_i, b_i$. We then fit the data and compute the predictive mean from equation \eqref{eqn:predictive_gpr} for a set of $1000$ randomly chosen test covariates $X_*$ with test features $\Phi_*$. The ground truth $y_*$ is chosen to be the predictive mean using all $N=10^4$ data.\footnote{Note that this is different to the predictive mean using an exact GP, but suffices for illustration purposes here, as the MSE is zero by construction when all data is used . }

\begin{figure}
\centering
\includegraphics{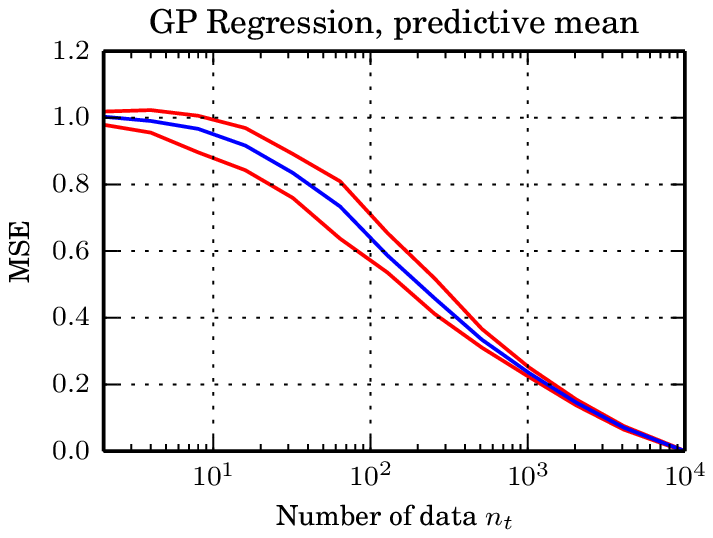}
\includegraphics{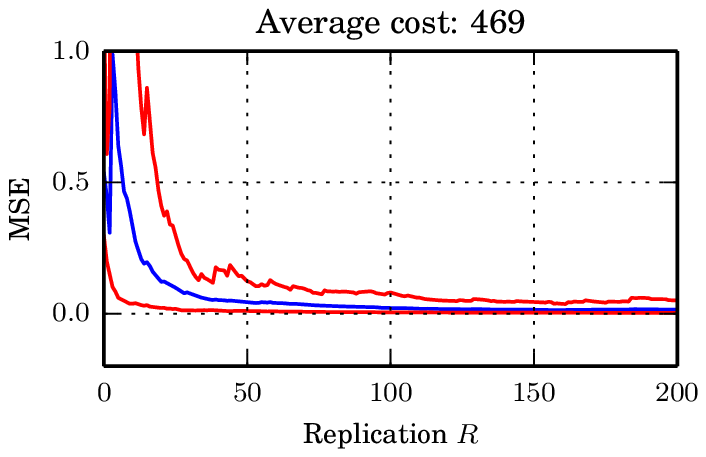}
\caption{\textbf{Top:} Convergence of MSE on test features as a function of data used for training. We randomly subsample available data, train the GP and compute the test MSE. 95\% error bars obtained by averaging over 200 trials. Note that the MSE eventually vanishes as we compare against the predictive mean obtained via using all data. \textbf{Bottom:} MSE as a function of debiasing replications. We compute multiple debiasing estimates for a given average cost and plot their running mean. The given average cost corresponds to a vertical slice in the above plot, i.e. predictions of constant sized mini-batches of size $469$ gives an MSE of more than $0.2$ while debiasing with the same average cost reaches almost zero error.}
\label{fig:debiasing_random_feature_regression}
\end{figure}

As before, we begin by exploring convergence of the desired posterior statistic, here averaged for \emph{multiple} test features  $\Phi_*$. For a given partial posterior size, we repeatedly sub-sample observations and compute the predictive mean. Figure \ref{fig:debiasing_random_feature_regression} (top) shows convergence of the mean squared error (MSE) of the predictive means for all test features as a function of partial posterior size. The MSE only gets close to zero when almost all data is used.  This is unlike in previous examples and therefore shows that the functional corresponding to GP regression, i.e.\ equation  \eqref{eq:gp_predictive_posterior_dual}, is more complicated.

We apply the debiasing scheme and compute the average computational complexity, i.e. the average size of all partial posteriors of a single debiasing replication. Given that complexity, Figure \ref{fig:debiasing_random_feature_regression} (top) shows the MSE if we were to average predictions over multiple mini-batches. In contrast, Figure \ref{fig:debiasing_random_feature_regression} (bottom) reveals that debiasing achieves a much better MSE \emph{at the same average computational cost}.

To our knowledge, none of the other approximate or exact sub-sampling-based MCMC schemes can be applied to this example. We therefore resort to comparing against a popular approximate inference method for such GP models. 

\subsection{Comparing to stochastic variational inference on real-world data}
Another way to approach Gaussian Processes in the Big Data context is via stochastic variational inference (GP-SVI). \citet{Hensman2013} combine a decade's work on sparse GPs, variational bounds, and stochastic gradient descent to fit huge GP models in a streaming fashion. The clever usage of a number of approximations allows them to cut the computational costs from  $\mathcal{O}(N^3)$ down to $\mathcal{O}(m^3)$, where $m$ is the number of inducing variables and constants depending on number of iterations and mini-batch size.

\paragraph{Airtime delays}
We apply a combination of random Fourier features and debiasing (as in previous Section) to the real-world problem of predicting arrival time delays in flight records \citep[Section 4.3]{Hensman2013}. This involves $N=700,000$ data consisting of 8-dimensional covariates and real labels. We aim to estimate the predictive mean of a GP for $100,000$ randomly chosen test covariates. We use the exponentiated quadratic covariance function, with a finite-rank expansion via random Fourier features. For the sake of simplicity, we do not apply a different length-scale to each dimension and include no bias term. Instead, we centre the data and re-scale to unit variance in a preprocessing step\footnote{Indeed, we were not able to obtain significant differences working with varying length-scales or other hyperparameters. Furthermore, \citet{Hensman2013} do not report predictive variance, for which tuning such parameters is more essential. However, random Fourier features are easily adapted to such covariance functions.}. We match the number of random Fourier features $m=1000$ to the number of inducing points in the GP-SVI experiment.

In debiasing,  We use the minimum batch size $a=500$, and set the trucation distribution to match an average computational cost of roughly $2773$ for each of the $R=100$ replications, which is an order of magnitude less than the batch size of $1000$ for $1000$ iterations in the GP-SVI experiment.

\begin{figure}
\centering
\includegraphics{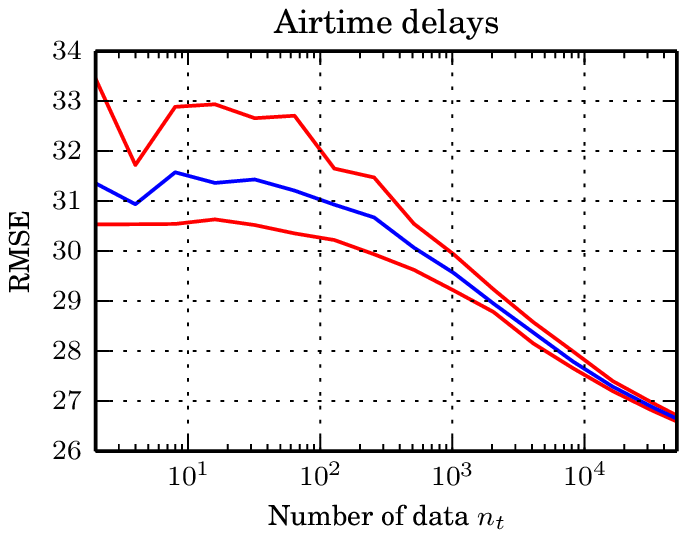}
\includegraphics{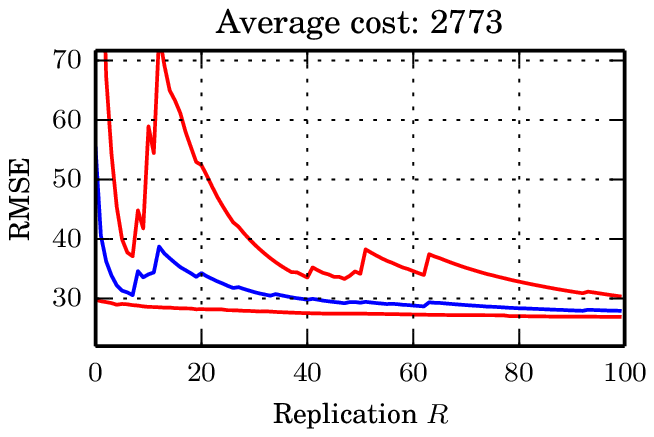}
\caption{Debiasing for the airtime delays dataset. \textbf{Top:} Convergence of RMSE as a function of mini-batch size using random Fourier features. The results using $1000$ data are slightly better to those reported in \citet{Hensman2013}, speaking for our choice of hyper-parameters. \textbf{Bottom:} With an order of magnitude less average computational cost than GP-SVI (see text), we are able to reproduce a comparable RMSE on $10^5$ randomly chosen test covariates. In particular, as in the previous GP example, Figure \ref{fig:debiasing_random_feature_regression}, we are able to obtain a better RMSE than averaging predictions obtained from constant batch sizes with the same computational costs. 95$\%$ error bars are computed over 20 repetitions.}
\label{fig:gp_airlines_debiasing}
\end{figure}

Remarkably, as shown in Figure \ref{fig:gp_airlines_debiasing}, debiasing outperforms \citet[Figure 7]{Hensman2013}. GP-SVI achieves a square rooted mean squared error (RMSE) of $32.6$, while debiasing achieves less than $27.9$. Care has to be taken when concluding from these RMSE comparisons -- both methods are likely to be improved by tuning, the full experimental protocols are not available, there are slight differences in finite-rank approximations, etc. Instead, we make the point that debiasing achieves a \emph{competitive} performance. However, while GP-SVI is highly engineered to these very same GP regression models, debiasing is a more general method for estimation in Bayesian inference -- with GP regression only being one of its applications. 

While this example is promising, we leave a thorough comparison with streaming variational Bayes for future work.

\subsection{Reducing bias in streaming applications}
The debiasing formalism is easily applicable in a scenario where the amount of data is unknown or unlimited, e.g.
in a streaming scenario. Previously, we discussed targeting the posterior given a fixed number $N$ of observations $\{x_i\}_{i=1}^N$, and
constructing the stochastic truncation variable $T$ such that a small probability remains that 
\emph{all observations} are used for computing the desired expectation.
In contrast, in the streaming scenario, we are unable to process all observations at a time, and the nature of the problem forces us to
process observations in batches -- which are discarded afterwards.
Debiasing is still possible: we fix a worst case budget $N_\text{max}$, which is the largest number of observations that can be processed at a time (e.g. guided by the hardware restrictions).
$N_\text{max}$ then replaces $N$ in the static case: the stochastic truncation variable $T$ allows processing $N_\text{max}$ observations at maximum.
This means that the bias with respect to the full posterior (of unknown size) still remains. However, as $R\ll N_\text{max}$, it is typically of the order $\mathcal{O}\left(1/\sqrt{N_\text{max}}\right)$,
and is therefore subsumed by the error bars over $R$ replications, which are of the order $\mathcal{O}(1/\sqrt{R} )$.

Note that in the streaming scenario, no fully unbiased scheme is available.

\begin{figure}
\centering
\includegraphics{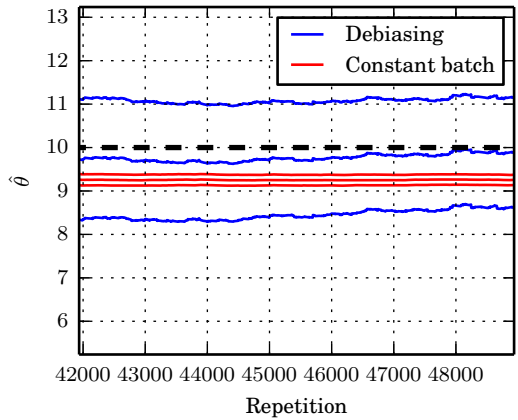}
\caption{Reduced bias for mean estimation in a Gaussian model with known variance. Our estimator is less biased and has more appropriate error bars than averaging over constant batch sizes, which is overconfidently biased. Both approaches have the same average computational costs.}
\label{fig:streaming_mean}
\end{figure}

\paragraph{Toy example}
We compare the debiasing scheme with the constant-batch scheme on a simple posterior mean estimation
in a Gaussian model with known variance, where $x_i\sim \mathcal N(\theta,5000^2)$ with prior $\theta\sim\mathcal N(0,50^2)$ and true $\theta=10$. Results are given in Figure \ref{fig:streaming_mean}. They show that the debiasing estimator is less biased and has more appropriate error bars where the constant batch-size approach is overconfident and strongly biased.
The constant batch size was chosen to make the computational cost of the two schemes comparable. Results show estimates towards the end of the total of $50,000$ replications after each scheme has streamed around $10^9$ datapoints. 

\section{Discussion}
\label{sec:discussion}

In this section, we present shortcomings and problems, both conceptual experimental, that have to be addressed in future work. We close by summarising our contributions.

\paragraph{Bias from memory restrictions}
In order for the estimator in \eqref{eq:telesoping_estimator}  to be unbiased, one needs to assign a non-zero probability for \emph{each} of the possible values of the truncation variable $T$ -- and the resulting partial posterior expectations need to be \emph{computable in finite time}. In all presented examples, sampling large values of $T$ only results in a long runtime. However, such large $T$ might also result in a partial posterior statistics that are impossible to estimate due to restrictions of available computing resources. A common example are memory limits arising from large Gaussian covariance matrices, see for example \citep{Lyne2013}. We side-stepped such problems in our experiments by using a finite-rank kernel expansion. However, in general our estimator is not unbiased in cases where partial posteriors exceed available machine memory. In practice, allowing a fixed computational budget and tweaking the truncation distribution such that larger values are almost never sampled, yields good results. Developing a more sophisticated solution is left for future work.

\paragraph{Convergence on short posterior paths}
In experiments on smaller datasets, we could not beat full posterior sampling in terms of estimation error per computation time. Only when the sub-linear average computational complexity $\mathcal{O}(N^{1-\alpha})$ is \emph{significantly less} than $\mathcal{O}(N)$, debiasing outperforms MCMC. It will be interesting to study the connection of data size $N$ and truncation parameter $\alpha$ for different classes of posterior functionals $\varphi$.

\begin{figure}
\centering
\includegraphics{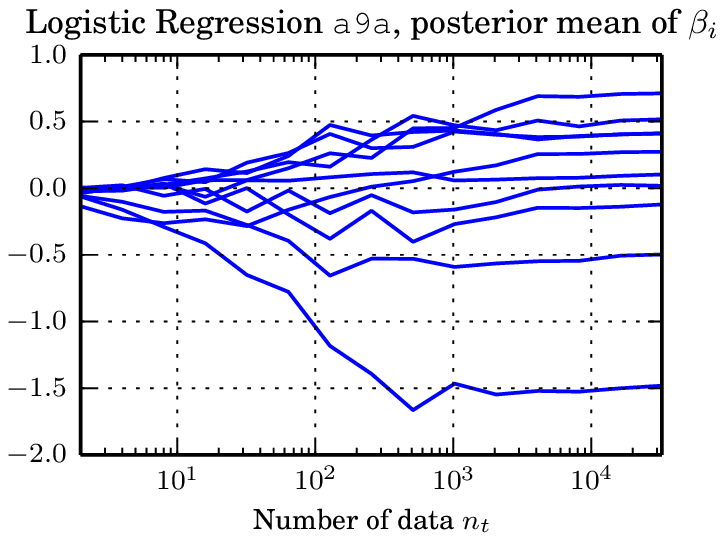}
\includegraphics{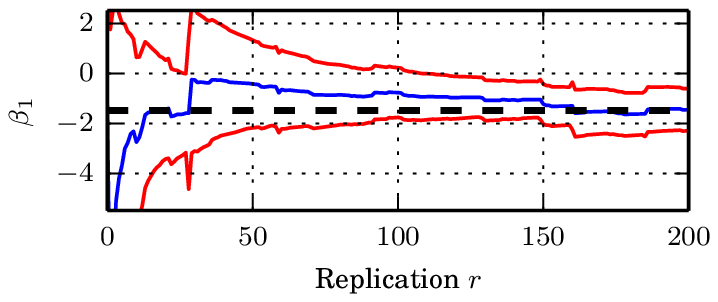}
\caption{Logistic regression on the \texttt{a9a} dataset consisting of $N=32561$ data of dimension $D=123$. We estimate posterior mean of the first regression weight $\beta_1$ for increasing data size. \textbf{Top:} Convergence of partial posterior statistics. \textbf{Bottom:} Convergence of the debiasing estimator looks promising. However, at around iteration 100, full posterior sampling is performed.
 }
\label{fig:logistic_regression_negative}
\end{figure}

Figure \ref{fig:logistic_regression_negative} shows results for (sparse) logistic regression  on the \texttt{a9a} dataset \citep{Lin2008,Welling2011}, which consists of $N=32561$ covariates of dimension $123$. Using the same model as in Section \ref{sec:logistic_toy_experiment}, we aim to estimate the posterior mean of the first regression weight $\beta_1$. Note that full posterior sampling on this dataset takes \emph{days}. Figure \ref{fig:logistic_regression_negative} (top) shows convergence of partial posterior statistics, which tend to stabilise from about 1000 data. Convergence of debiasing, Figure \ref{fig:logistic_regression_negative} (middle), behaves well at first sight. However, as $N$ in this case is relatively small, the probability to sample a partial posterior path truncation that includes the \emph{whole} dataset is relatively high. In the presented debiasing run, this happens around replication 100. As this results in \emph{full} posterior sampling, debiasing is pointless. Unfortunately, the convergence at this point has not yet reached an acceptable level.

\subsection*{Summary}
We presented an alternative perspective on large-scale Bayesian inference problems, and developed a novel framework for approaching those in practice. For cases where the goal is \emph{estimation} of Bayesian posterior expectations, rather than \emph{simulation} from the posterior,  we side-stepped the many serious convergence problems arising from employing approximate transition kernels of Markov chains for simulation. By exploiting the debiasing Lemma, we were able to estimate these posterior statistics efficiently from partial posterior statistics. Data complexity is sub-linear in $N$, no bias is introduced, variance is finite.

Implementing our approach is trivial as it exploits existing work on MCMC and easily fits in with other inference schemes. Free parameters are easy to tune. It furthermore is embarrassingly parallelisable. We conducted experiments to illustrate cases where debiasing can accurately and confidently estimate posterior statistics before competing simulation methods are able to produce a single estimate. The presented methodology is not limited to factorising likelihoods or  MCMC as an internal inference scheme. We carried out experimental examples that showcased competitiveness of debiasing compared to full posterior sampling and stochastic variational inference.

Most essential areas for future work are (i) exploring the computation-variance tradeoff in detail, also in context of \emph{other} than geometric truncation distributions (ii) dealing with finite time bias when MCMC is used, (iii) a thorough formal and experimental comparison with other large-scale inference schemes such as stochastic variational inference.

\bibliography{biblio}

\newpage
\appendix
\onecolumn
\section{Computational complexity and variance for geometric batch schedule}
\label{sec:Appendix_complexity_variance}
In this section, we show that for the simple choice of a geometrically increasing batch schedule, and a geometric stochastic truncation variable, it is possible to obtain sub-linear expected complexity of the debiasing scheme. Furthermore, variance remains bounded by a constant as $N$ increases. 

\paragraph{Number of likelihood evaluations}
For simplicity, we assume the common ratio $2$, i.e.\ the batch sizes are $n_{1}=a,\dots,n_{L}=2^{L-1}a=N$, where $L=\log_{2}\left(N/a\right)+1$, and $a$ is the smallest batch size considered, and $\log_2(N/a)$ is an integer. We express computational costs in terms of the number of likelihood evaluations $\mathcal L$. It is easy to see that $\mathcal L$ is a function of the stochastic truncation variable $T$, i.e.
\begin{eqnarray*}
\mathcal L(T) = M\sum_{t=1}^{T}n_{t} & = & Ma\left(2^{T}-1\right),
\end{eqnarray*}
where $M$ is the length of the MCMC chains (assumed to be constant throughout). Namely, $T$ chains need to be run on partial posteriors w.r.t.\ $n_1,\dots,n_T$ datapoints respectively.

We write the truncation probability as $p_{t}:=\mathbb P(T=t)\propto2^{-\alpha t}$ for some $\alpha\in(0,1)$. The normalizing constant of the corresponding density is given by $Z_{\alpha}=\sum_{t=1}^{L}2^{-\alpha t}=2^{-\alpha}\left(1+2^{-\alpha}+\dots2^{-\alpha(L-1)}\right)=2^{-\alpha}\frac{1-2^{-\alpha L}}{1-2^{-\alpha}}\approx\frac{2^{-\alpha}}{1-2^{-\alpha}}$, leading to the expected number of likelihood evaluations being
\begin{eqnarray*}
\mathbb E\left[\mathcal L(T)\right]& = & \frac{Ma}{Z_{\alpha}}\sum_{t=1}^{L}\left(2^{t}-1\right)2^{-\alpha t}\\
& = & \frac{Ma}{Z_{\alpha}}\left(\sum_{t=1}^{L}2^{(1-\alpha)t}-Z_{\alpha}\right)\\
& = & \frac{Ma}{Z_{\alpha}}2^{1-\alpha}\frac{2^{\left(1-\alpha\right)L}-1}{2^{1-\alpha}-1}-a\\
& \approx & 2Ma\frac{1-2^{-\alpha}}{1-2^{\alpha-1}}\left(N/a\right)^{1-\alpha}\\
& = & O\left(Ma(N/a)^{1-\alpha}\right),
\end{eqnarray*}
i.e. the overall complexity is sub-linear in the number of observations for $\alpha>0$.

\paragraph{Variance}
The tail of $T$ is
\begin{eqnarray*}
\mathbb{P}\left[T\geq t\right] & = & \frac{1}{Z_{\alpha}}2^{-\alpha t}\left(1+2^{-\alpha}+\dots2^{-\alpha(L-t)}\right)\\
 & = & \frac{1}{Z_{\alpha}}2^{-\alpha t}\frac{1-2^{-\alpha(L-t+1)}}{1-2^{-\alpha}}\\
 & = & 2^{-\alpha(t-1)}\frac{1-2^{-\alpha(L-t+1)}}{1-2^{-\alpha L}}\\
 & = & \frac{2^{-\alpha(t-1)}-2^{-\alpha L}}{1-2^{-\alpha L}}.
\end{eqnarray*}
In addition to $\alpha$, variance will also depend on the rate of convergence of partial posterior expectations to the full posterior expectation. We denote the difference between the expectation estimated on a partial posterior $\pi_t$ (corresponding to $n_t$ points) and the expectation estimated on the full posterior by
\begin{align*}
\delta_t:=\hat{\mathbb E}_{\pi_t}\{\varphi(\theta)\}-\hat{\mathbb E}_{\pi_{N}}\{\varphi(\theta)\}.
\end{align*}
Note that we have $\delta_t=0$ almost surely for $t>L$ as the sequence of estimators terminates with the full posterior. This suffices that for any finite $N$, variance of the debiasing scheme is finite as well. However, it might grow without bound as $N$ grows large -- which is an undesirable situation. Remarkably, it is possible to ensure that variance is $\mathcal{O}_N(1)$, i.e. it  \emph{stays bounded as $N$ increases}. Namely, let us assume that for large enough $N$, there exist a constant $c$ and $\beta>0$, such that $\forall t\leq L$:
\begin{align}
\label{eqn:beta_equation}
 \mathbb{E}\left\{\left|\delta_t\right|^{2}\right\}\leq \frac{c}{n^\beta_{t}}=\frac{c}{a^{\beta}2^{\beta(t-1)}}.
\end{align}
The coefficient $\beta$ clearly depends on the function $\varphi$: it is typically $1$ for simple models and $\varphi(\theta)=\theta$ and could be closer to zero for complicated functionals $\varphi$ exhibiting slow convergence of partial posterior expectations. Now, from Lemma \ref{eq:telesoping_estimator}, the second moment of the debiasing estimator is precisely
\begin{eqnarray*}
\mathbb E \left\{\left(\phi_{T}^{*}\right)^2\right\}&=&\sum_{t=1}^{L}\frac{\mathbb{E}\{\left| \delta_{t-1}\right|^{2}\}-\mathbb{E}\{\left| \delta_t\right|^{2}\}}{\mathbb{P}\left[T\geq t\right]}\\
&\leq&\sum_{t=1}^{L}\frac{\mathbb{E}\{\left| \delta_{t-1}\right|^{2}\}}{\mathbb{P}\left[T\geq t\right]}\\
&\leq&\frac{c2^\beta\left(1-2^{-\alpha L}\right)}{a^\beta}\sum_{t=1}^{L}\frac{1}{2^{(\beta-\alpha)\left(t-1\right)}-2^{\beta(t-1)-\alpha L}},
\end{eqnarray*}
where the last sum remains finite for $L\to\infty$ as long as $\alpha<\beta$. This implies that the variance of the scheme remains bounded by a constant as the number of observations $N$ grows large. In terms of $\alpha$ and $\beta$, this upper bound on the second moment is approximately $\frac{c2^{\beta}}{a^\beta(1-2^{\alpha-\beta})}$ for large $L$.

\begin{figure}
\centering
\includegraphics{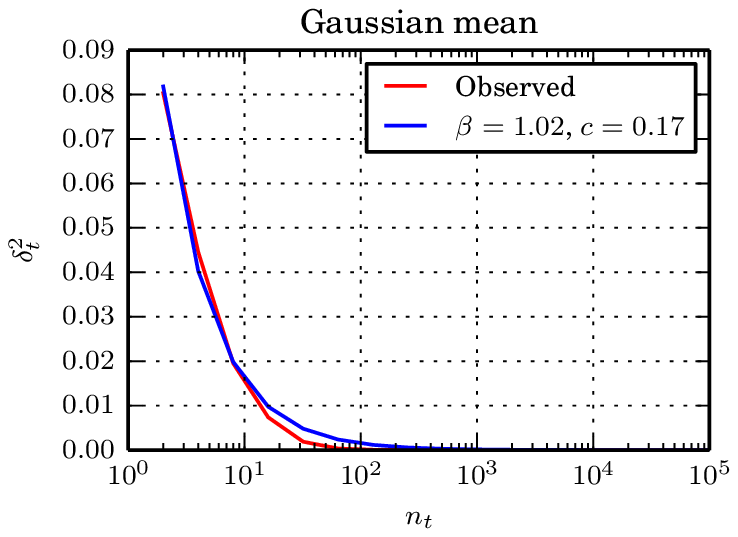}
\includegraphics{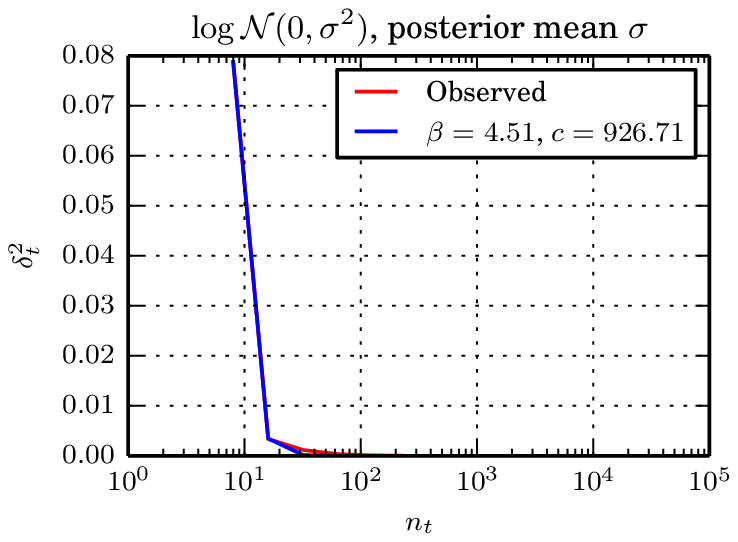}
\includegraphics{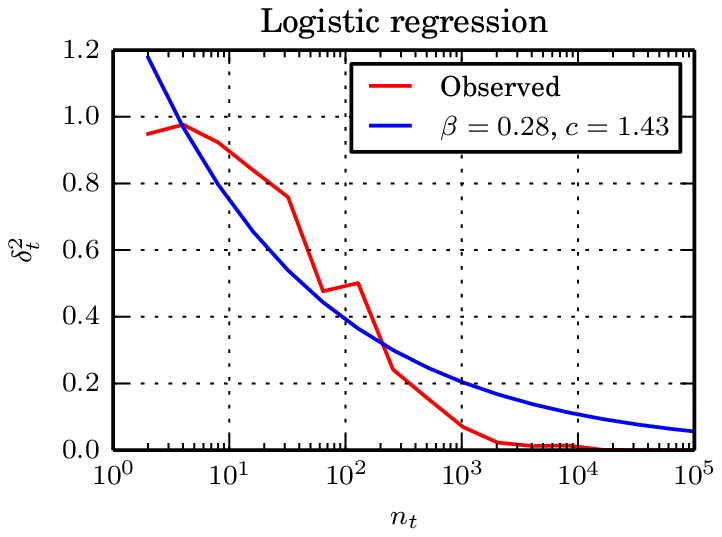}
\includegraphics{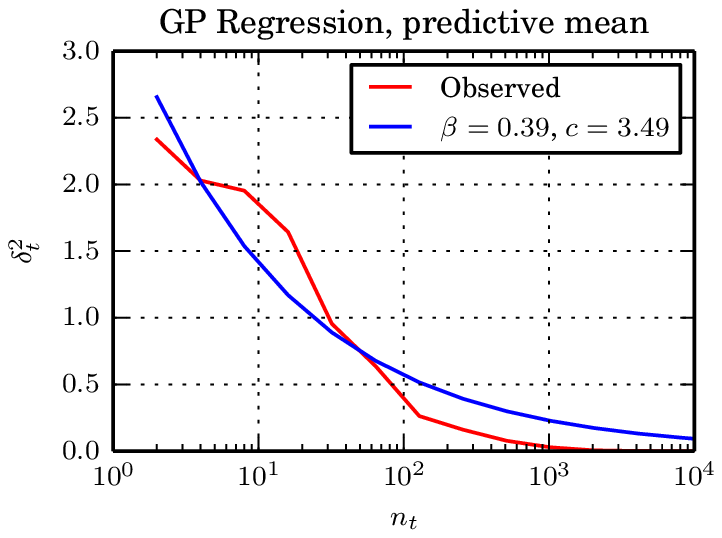}
\caption{Empirical least squares fits of $cn_t^{-\beta}$ in equation \eqref{eqn:beta_equation} to observed partial posterior statistics' convergence. \textbf{Top-left:} Gaussian mean, from Figure \ref{fig:posterior_path}. \textbf{Top-right:} Log-Gaussian standard deviation, from Figure \ref{fig:bardenet_convergence}. \textbf{Bottom-left:} First regression weight of synthetic logistic regression, from Figure \ref{fig:logistic_toy}. \textbf{Bottom-right:} A single prediction in approximate GP regression, from Figure \ref{fig:debiasing_random_feature_regression}. }
\label{fig:beta_fits}
\end{figure}

Figure \ref{fig:beta_fits} shows fits of equation \eqref{eqn:beta_equation} to the convergence rates of the $\delta_t^2$ for various of the presented examples. This shows that $\beta$ can be chosen close to $1$ or larger for simple, quickly converging posterior statistics. Note that values larger than 1 imply a constant average computational cost of debiasing (independently of the number of observations). However, note that these are empirical fits, sensitive to noise etc. In practice, it is possible to estimate $\beta$ by investigating the desired expectations on the first few partial posteriors only -- comparisons being w.r.t. the expectation given the largest batch considered among these, rather than w.r.t. the full posterior. We stress that getting an accurate estimate of $\beta$ is not required for our scheme - a conservative lower bound on $\beta$ suffices to ensure that the parameter $\alpha$ used in the truncation probabilities is smaller and thus variance remains bounded.

\paragraph{Minimising asymptotic variance} From \eqref{eq:clt_work_variance}, the variance $\times$ cost determines the asymptotic variance of the debiasing scheme. Thus, provided $\beta$ is known, one can use the derivations above to select the parameter $\alpha$ in the stochastic truncation distribution,
\[
\alpha=\underset{\alpha'\in(0,\beta)}{\text{argmax}}\frac{a^{\alpha'}\left(1-2^{-\alpha'}\right)}{(1-2^{\alpha'-1})(1-2^{\alpha'-\beta})}N^{1-\alpha'}.
\]

\end{document}